\documentclass[conference]{IEEEtran}
\IEEEoverridecommandlockouts
\usepackage{cite}
\usepackage{amsmath,amssymb,amsfonts}
\usepackage{algorithmic}
\usepackage{graphicx}
\usepackage{textcomp}
\usepackage{xcolor}
\usepackage{booktabs}
\usepackage{multirow}
\usepackage{siunitx}
\usepackage{threeparttable}
\usepackage{array}

\newcolumntype{C}[1]{>{\centering\arraybackslash}p{#1}}
\newcolumntype{R}[1]{>{\raggedleft\arraybackslash}p{#1}}

\def\BibTeX{{\rm B\kern-.05em{\sc i\kern-.025em b}\kern-.08em
    T\kern-.1667em\lower.7ex\hbox{E}\kern-.125emX}}
\begin{document}

\title{PATS: Proficiency-Aware Temporal Sampling for Multi-View Sports Skill Assessment}

\author{\IEEEauthorblockN{1\textsuperscript{st} Edoardo Bianchi}
\IEEEauthorblockA{\textit{Faculty of Engineering} \\
\textit{Free University of Bozen-Bolzano}\\
Bozen-Bolzano, Italy \\
edbianchi@unibz.it}
\and
\IEEEauthorblockN{2\textsuperscript{nd} Antonio Liotta}
\IEEEauthorblockA{\textit{Faculty of Engineering} \\
\textit{Free University of Bozen-Bolzano}\\
Bozen-Bolzano, Italy \\
antonio.liotta@unibz.it}
}

\maketitle

\begin{abstract}
Automated sports skill assessment requires capturing fundamental movement patterns that distinguish expert from novice performance, yet current video sampling methods disrupt the temporal continuity essential for proficiency evaluation. To this end, we introduce Proficiency-Aware Temporal Sampling (PATS), a novel sampling strategy that preserves complete fundamental movements within continuous temporal segments for multi-view skill assessment. PATS adaptively segments videos to ensure each analyzed portion contains full execution of critical performance components, repeating this process across multiple segments to maximize information coverage while maintaining temporal coherence. Evaluated on the EgoExo4D benchmark with SkillFormer, PATS surpasses the state-of-the-art accuracy across all viewing configurations (+0.65\% to +3.05\%) and delivers substantial gains in challenging domains (+26.22\% bouldering, +2.39\% music, +1.13\% basketball). Systematic analysis reveals that PATS successfully adapts to diverse activity characteristics—from high-frequency sampling for dynamic sports to fine-grained segmentation for sequential skills—demonstrating its effectiveness as an adaptive approach to temporal sampling that advances automated skill assessment for real-world applications. Project page: https://edowhite.github.io/PATS
\end{abstract}


\section{Introduction}
Automated sports skill assessment represents a critical challenge with applications in training, coaching, and talent development. Unlike action recognition, skill assessment requires capturing how well an action is executed through subtle temporal dynamics that distinguish expert from novice performance.

In sports contexts, expertise manifests through temporal patterns: the rhythm of basketball dribbling, timing coordination in soccer control, or fluid progression in athletic techniques. These patterns emerge from the execution of fundamental movement components—the basic building blocks of skilled performance that must be observed in their natural temporal context to accurately assess proficiency.

However, current video sampling approaches fundamentally disrupt this continuity. Uniform sampling at regular intervals potentially misses critical transitions, random sampling lacks systematic coverage, and motion-density sampling ignores the structured temporal nature of skilled performance. Even segment-based approaches typically employ sparse sampling within segments, breaking the natural flow essential for skill assessment. These limitations stem from a failure to recognize that athletic proficiency manifests through structured temporal patterns requiring continuous observation of complete fundamental movements.

Our key insight is that effective skill assessment requires analyzing continuous video portions containing at least one complete fundamental movement, repeated across multiple segments to maximize information capture while preserving temporal coherence.

In this work, we introduce Proficiency-Aware Temporal Sampling (PATS), a novel sampling strategy specifically designed for multi-view sports skill assessment. PATS preserves complete fundamental movements within continuous temporal segments, ensuring that each analyzed portion contains the full execution of critical performance components. By repeating this process across multiple non-overlapping segments, PATS maximizes information coverage while maintaining the temporal continuity essential for accurate proficiency assessment.

This work presents the following key contributions:
\begin{itemize}
\item A proficiency-aware sampling strategy that preserves fundamental movements within continuous temporal segments.
\item Synchronized multi-view integration that preserves temporal coherence across camera perspectives.
\item Architecture-agnostic design enabling seamless integration with existing temporal modeling frameworks without computational overhead.
\end{itemize}

When applied to SkillFormer \cite{skillformer}, a recent and competitive multi-view proficiency estimation framework, PATS achieves consistent improvements across all viewing configurations (+0.65\% to +3.05\%) and delivers substantial scenario-specific gains (+26.22\% bouldering, +2.39\% music, +1.13\% basketball), establishing new accuracy standards and demonstrating its ability to capture coherent and meaningful temporal patterns essential for skill assessment.

PATS operates as a preprocessing step that enhances model accuracy without adding computational overhead, maintaining efficiency while providing an adaptive approach to temporal sampling for sports skill assessment.

\section{Background and Related Work}
\subsection{Temporal Sampling Methods}
Temporal sampling determines which frames to extract from video sequences for processing. Traditional approaches have evolved from uniform methods to sophisticated content-aware strategies. While Temporal Segment Networks (TSN)~\cite{wang2019temporal} achieve computational efficiency through sparse sampling (selecting few frames distributed across the video duration), they sacrifice temporal continuity essential for skill assessment. Recent advances including motion-density methods, temporal correspondence techniques like sandwich sampling~\cite{liu2025future}, channel sampling strategies~\cite{kim2022capturing}, and temporal contextualization~\cite{kim2024leveraging} have primarily focused on general action recognition, with limited consideration for the specialized requirements of proficiency assessment where temporal dynamics distinguish expert from novice performance.

\subsection{Action Quality Assessment}
Action Quality Assessment (AQA) evaluates how well actions are performed, requiring sensitivity to subtle performance differences~\cite{zhou2024comprehensive}. Modern methods have evolved from handcrafted features to deep learning architectures using pre-trained backbones like I3D \cite{carreira2017quo} and transformers \cite{liu2022video}. Recent advances include spatial-aware modeling \cite{wang2021tsa}, temporal-aware approaches \cite{xu2022finediving}, and quality prediction strategies \cite{tang2020uncertainty, yu2021group}.

Multi-modal approaches integrate skeleton features~\cite{gsp, skeletonAR} and audio information~\cite{skatinmixer, pamfn} for enhanced understanding. Multi-view methods show particular promise: SkillFormer~\cite{skillformer} processes synchronized egocentric and exocentric videos through a TimeSformer \cite{timesformer} backbone with LoRA \cite{lora} fine-tuning, using CrossViewFusion with multi-head cross-attention for view-specific feature fusion. However, it relies on uniform sampling, which fails to preserve complete temporal dynamics. EgoPulseFormer~\cite{egoppg} leverages physiological signals from eye-tracking cameras. Datasets such as EgoExoLearn~\cite{egoexolearn} and EgoExo4D~\cite{egoexo4d} provide multi-perspective benchmarks.

Existing methods inherit temporal sampling limitations, failing to preserve the temporal dynamics that distinguish expert from novice performance. Our work addresses these limitations through specialized sampling strategies designed for skill assessment scenarios.

\begin{figure*}[t]
    \centering
    \includegraphics[width=1\linewidth]{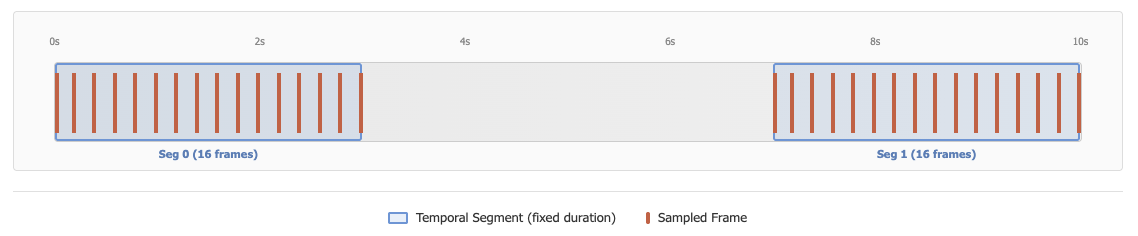}
    \caption{\label{fig:PATS}
    In this configuration, PATS extracts $N_{target} = 32$ frames from $N_s = 2$ continuous temporal segments of duration $d_s = 3 s$ from a 10 s video. Within each segment, $\lfloor N_{target}/N_s \rfloor = 16$ frames are sampled uniformly (red vertical lines), preserving temporal continuity within segments. Segment positioning with automatic spacing prevents overlap and ensure comprehensive temporal coverage. This configuration is used in the basketball and bouldering domains.}
\end{figure*}

\section{Proposed Methodology}
\label{sec:method}
\subsection{Proficiency-Aware Temporal Sampling (PATS)}
We introduce Proficiency-Aware Temporal Sampling (PATS), a novel temporal sampling strategy designed to preserve the sequential nature of skilled performances for accurate proficiency assessment. Unlike traditional uniform sampling methods that randomly select frames across a video, PATS extracts continuous temporal segments that maintain the natural flow of athletic movements and performance dynamics, as demonstrated in Figure \ref{fig:PATS}.

\subsubsection{Method Overview and Parameters}
PATS is controlled by three key parameters that balance temporal coverage with continuity preservation: $N_{target}$ (total number of frames to extract from the input video), $N_s$ (number of temporal segments to divide the video into), and $d_s$ (desired duration in seconds for each temporal segment).

The core principle of PATS is to extract $N_s$ continuous temporal segments of duration $d_s$, distributed across the entire video timeline, yielding exactly $N_{target}$ frames total. This approach ensures comprehensive temporal coverage while preserving the sequential relationships critical for skill assessment.

\subsubsection{Adaptive Segment Positioning}
Given an input video of total duration $T$ seconds with $N_{total}$ frames captured at frame rate $fps$, PATS first determines the effective segment duration to prevent temporal overlap and ensure adequate spacing:
\begin{equation}
d_{s,eff} = \min\left(d_s, \frac{0.8T}{N_s}\right)
\end{equation}
The 0.8 scaling factor limits each segment to 80\% of its theoretical maximum duration, creating natural spacing between segments and preventing temporal overlap. 

Segment start times are then distributed uniformly across the feasible temporal range. To ensure segments remain within video boundaries, the maximum allowable start time is:
\begin{equation}
t_{max} = \max(0, T - d_{s,eff})
\end{equation}

For multiple segments ($N_s > 1$), start times are positioned using:
\begin{equation}
t_{start,i} = i \times \frac{t_{max}}{N_s - 1}, \quad i = 0, 1, ..., N_s-1
\end{equation}

For single-segment scenarios ($N_s = 1$), the segment begins at $t_{start,0} = 0$.

\subsubsection{Frame Allocation and Extraction}
The algorithm ensures exactly $N_{target}$ frames are extracted by distributing them across segments. The number of frames allocated to each segment is:
\begin{equation}
N_{frs,i} = \left\lfloor \frac{N_{target}}{N_s} \right\rfloor + \mathbf{1}[i < N_{target} \bmod N_s]
\end{equation}
where $\mathbf{1}[\cdot]$ is the indicator function. This allocation guarantees that remainder frames are distributed to the first segments, ensuring exactly $N_{target}$ total frames.

For each segment $i$ with temporal boundaries $[t_{start,i}, t_{start,i} + d_{s,eff}]$, the corresponding frame indices are computed as:
\begin{align}
f_{s,i} &= \lfloor t_{start,i} \times fps \rfloor \\
f_{e,i} &= \min(\lfloor (t_{start,i} + d_{s,eff}) \times fps \rfloor, N_{total})
\end{align}

Within each segment, frames are extracted using continuous sampling:
\begin{equation}
\text{frames}_i = \begin{cases}
\left\lfloor \frac{f_{s,i} + f_{e,i}}{2} \right\rfloor & \text{if } N_{frs,i} = 1 \\
\text{linspace}(f_{s,i}, f_{e,i} - 1, N_{frs,i}) & \text{otherwise}
\end{cases}
\end{equation}

\subsubsection{Robust Edge Case Handling}
The algorithm includes comprehensive handling for challenging scenarios to ensure reliable operation across diverse video conditions:

\textit{Insufficient Video Duration}: If $T \leq 0$ or $N_{total} < N_{target}$, the method falls back to uniform sampling.

\textit{Minimal Segment Duration}: When $d_{s,eff} < 0.5$ seconds, the algorithm reverts to uniform sampling to preserve temporal coherence.

\textit{Boundary Violations}: Segment boundaries exceeding video limits are automatically adjusted to ensure $f_{e,i} \leq N_{total}$.

\textit{Frame Count Adjustment}: The algorithm guarantees exactly $N_{target}$ output frames through uniform subsampling (excess frames) or cyclic repetition (insufficient frames).

The final output is a sorted list of exactly $N_{target}$ frame indices clipped to $[0, N_{total} - 1]$, ensuring robust behavior across all input conditions.

\subsection{Integration with SkillFormer}
We integrate PATS with SkillFormer~\cite{skillformer} as it represents the current state-of-the-art for proficiency estimation on EgoExo4D \cite{egoexo4d}, with both approaches targeting sports domains where temporal continuity distinguishes expert from novice performance.

PATS enhances SkillFormer's multi-view architecture by preserving temporal continuity essential for cross-view fusion. The CrossViewFusion module relies on meaningful temporal relationships to capture performance nuances across camera perspectives. By providing continuous temporal segments rather than sparse frames, PATS enables more effective attention computations for multi-view movement analysis.

PATS integrates seamlessly into SkillFormer, replacing the original uniform sampling strategy while maintaining full architectural compatibility and enhancing proficiency assessment without structural modifications.

\begin{table*}[ht]
\centering
\caption{Comparison with EgoExo4D proficiency estimation baselines (from \cite{egoexo4d}) and SkillFormer (from \cite{skillformer}). We report accuracy (\%) for egocentric (Ego), exocentric (Exos), and combined views (Ego+Exos). SkillFormer + PATS outperforms the baselines in all settings. Bold denotes the best accuracy; underlined values indicate second-best.}
\footnotesize
\begin{tabular}{lcccccc}
\toprule
\textbf{Method} & \textbf{Pretrain} & \textbf{Ego} & \textbf{Exos} & \textbf{Ego+Exos} & \textbf{Params} & \textbf{Epochs} \\
\midrule
Random & - & 24.9 & 24.9 & 24.9 & - & - \\
Majority-class & - & 31.1 & 31.1 & 31.1 & - & - \\
TimeSformer & - & 42.3 & 40.1 & 40.8 & 121M & 15 \\
TimeSformer & K400 & 42.9 & 39.1 & 38.6 & 121M & 15 \\
TimeSformer & HowTo100M & \underline{46.8} & 38.2 & 39.7 & 121M & 15 \\
TimeSformer & EgoVLP & 44.4 & 40.6 & 39.5 & 121M & 15 \\
TimeSformer & EgoVLPv2 & 45.9 & 38.0 & 37.8 & 121M & 15 \\
\midrule
SkillFormer-Ego & K600 & 45.9 & - & - & 14M & 4 \\
SkillFormer-Exos & K600 & - & \underline{46.3} & - & 20M & 4 \\
SkillFormer-EgoExos & K600 & - & - & \underline{47.5} & 27M & 4 \\
\midrule
\textbf{SkillFormer-Ego+PATS} & K600 & \textbf{47.3} & - & - & 14M & 4 \\
\textbf{SkillFormer-Exos+PATS} & K600 & - & \textbf{46.6} & - & 20M & 4 \\
\textbf{SkillFormer-EgoExos+PATS} & K600 & - & - & \textbf{48.0} & 27M & 4 \\
\bottomrule
\end{tabular}
\label{tab:baseline_comparison}
\end{table*}

\section{Experimental Setup}
\subsection{Dataset}
\label{subsec:dataset}
We evaluate on the Ego-Exo4D dataset \cite{egoexo4d}, which provides synchronized multi-view recordings across diverse real-world scenarios with over 1,200 hours of video from 740 participants. Our experiments focus on the official demonstrator proficiency benchmark, encompassing six activity domains: cooking, music, basketball, bouldering, soccer, and dance. Each video is annotated with four skill levels: \textit{Novice}, \textit{Early Expert}, \textit{Intermediate Expert}, and \textit{Late Expert}.

Following established protocols \cite{skillformer, egoppg}, we adopt the official dataset partitions, reserving 10\% of the training set for validation and conducting final evaluation on the official held-out validation set. 

\subsection{Implementation Details}
\label{subsec:implementation}
To ensure fair comparison, we maintain identical training configurations to the original SkillFormer implementation \cite{skillformer}. All models are initialized from a TimeSformer \cite{timesformer} backbone, pre-trained on Kinetics-600 \cite{carreira17}, and fine-tuned for 4 epochs using AdamW optimizer with weight decay of 0.01 and LoRA adaptation. We adopt SkillFormer's hyperparameters for each experimental setup.

The sole modification to the baseline framework is the temporal sampling strategy: while SkillFormer employs uniform frame sampling, our approach integrates PATS as described in Section~\ref{sec:method}. We systematically tune the PATS-specific parameters-total number of frames ($N_{target}$), number of temporal segments ($N_s$), and segment duration ($d_s$)-through grid search as detailed in Section~\ref{subsec:pats_params}, while keeping all other model and training configurations identical to the baseline.

Selected frames undergo standard preprocessing: resizing to 224-pixel shortest edge, center-cropping to $224 \times 224$, rescaling to $[0,1]$, and normalization with mean $[0.45, 0.45, 0.45]$ and standard deviation $[0.225, 0.225, 0.225]$.

Training was conducted on a single NVIDIA A100 GPU.

\subsection{PATS Hyperparameter Selection Rationale}
\label{subsec:pats_params}
We conduct a systematic grid search across PATS parameters to identify optimal configurations for different view setups and activity characteristics. Our parameter selection balances comprehensive evaluation with computational constraints while ensuring coverage of diverse temporal sampling scenarios.

We evaluate 24 and 32 frames per video to explore the trade-off between computational efficiency and temporal resolution. 

Segment count selection spans 2 to 12 segments: coarse segmentation (2 segments) captures preparation-execution dichotomy in continuous actions, medium segmentation (6-8 segments) enables multi-phase analysis for dynamic activities, and fine segmentation (12 segments) provides granular resolution for sequential tasks.

Duration configuration explores 1 s and 3 s segments based on sports biomechanics principles: short duration (1 s) captures rapid execution phases in high-frequency activities, while standard duration (3 s) encompasses complete movement sequences optimal for most scenarios.

The combination yields effective sampling rates from 0.89 to 5.33 FPS. High rates (4.0-5.33 FPS) prove optimal for dynamic activities requiring fine temporal resolution, while lower rates (0.89 FPS) suffice for structured, sequential activities like music performance.

\begin{table*}[ht]
\centering
\caption{All experimental configurations with training hyperparameters and overall accuracy (\%). Fixed across all runs: epochs=4, batch size=16, output dim=768, attention heads=16. Abbreviations: Segs=Segments, R=LoRA rank, A=LoRA alpha, Hid=Hidden dimension, Par=Parameters, Acc=Overall accuracy.}
\label{tab:all_configurations}
\footnotesize
\begin{tabular}{cccccccccccl}
\toprule
\textbf{Views} & \textbf{Frames} & \textbf{Segs} & \textbf{Duration (s)} & \textbf{FPS} & \textbf{R} & \textbf{A} & \textbf{Hid} & \textbf{LR} & \textbf{Par} & \textbf{Acc} & \textbf{Note} \\
\midrule
\multirow{4}{*}{Ego} & 24 & 6 & 3 & 1.33 & \multirow{4}{*}{32} & \multirow{4}{*}{64} & \multirow{4}{*}{1536} & \multirow{4}{*}{5e-5} & \multirow{4}{*}{14M} & 45.6 & \\
& 32 & 2 & 3 & 5.33 & & & & & & \underline{47.3} & Bouldering specialist \\
& 32 & 8 & 1 & 4.00 & & & & & & 42.0 & \\
& 32 & 12 & 3 & 0.89 & & & & & & 46.1 & Music specialist \\
\midrule
\multirow{3}{*}{Exos} & 24 & 6 & 3 & 1.33 & \multirow{3}{*}{48} & \multirow{3}{*}{96} & \multirow{3}{*}{2048} & \multirow{3}{*}{3e-5} & \multirow{3}{*}{20M} & 44.9 & \\
& 32 & 2 & 3 & 5.33 & & & & & & 43.4 & \\
& 32 & 8 & 1 & 4.00 & & & & & & 46.6 & Cooking specialist \\
& 32 & 12 & 3 & 0.89 & & & & & & 44.4 & \\
\midrule
\multirow{4}{*}{Ego+Exos} & 24 & 6 & 3 & 1.33 & \multirow{4}{*}{64} & \multirow{4}{*}{128} & \multirow{4}{*}{2560} & \multirow{4}{*}{2e-5} & \multirow{4}{*}{27M} & 44.6 & \\
& 32 & 2 & 3 & 5.33 & & & & & & 45.3 & Basketball specialist \\
& 32 & 8 & 1 & 4.00 & & & & & & 47.0 & Dancing specialist \\
& 32 & 12 & 3 & 0.89 & & & & & & \textbf{48.0} & \\
\bottomrule
\end{tabular}
\end{table*}

\begin{table*}[ht]
\centering
\caption{Optimal configuration patterns identified across different activity types.}
\label{tab:config_patterns}
\footnotesize
\begin{tabular}{lll}
\toprule
\textbf{Parameter} & \textbf{Optimal Range} & \textbf{Activity Type} \\
\midrule
Frame Count & 32 frames & All top-performing scenarios \\
\midrule
\multirow{2}{*}{Sampling Rate} & High FPS (4.0-5.33) & Dynamic activities (Basketball, Cooking, Dancing, Bouldering) \\
& Low FPS (0.89) & Structured activities (Music) \\
\midrule
\multirow{3}{*}{View Configuration} & Ego & Individual skill activities (Music, Bouldering) \\
& Exos only & External observation scenarios (Cooking) \\
& Ego+Exos & Team sports \& complex interactions (Basketball, Dancing) \\
\midrule
\multirow{3}{*}{Temporal Segmentation} & Few segments (2) & Continuous actions (Basketball, Bouldering) \\
& Medium segments (8) & Dynamic activities (Cooking, Dancing) \\
& Many segments (12) & Sequential, fine-grained tasks (Music) \\
\midrule
\multirow{2}{*}{Duration Strategy} & Short clips (1s) & High-frequency activities (Cooking, Dancing) \\
& Standard clips (3s) & Most scenarios (Basketball, Music, Bouldering) \\
\bottomrule
\end{tabular}
\end{table*}

\begin{table*}[ht]
\centering
\caption{Optimal configuration details for each scenario.}
\label{tab:best_configurations}
\footnotesize
\begin{tabular}{lccccccl}
\toprule
\textbf{Scenario} & \textbf{Best Acc. (\%)} & \textbf{Views} & \textbf{Frames} & \textbf{Segs} & \textbf{Duration (s)} & \textbf{FPS} & \textbf{Configuration Strategy} \\
\midrule
Basketball & \textbf{78.76} & Ego+Exos & 32 & 2 & 3 & 5.33 & Rapid sampling, minimal fragmentation \\
Cooking & \textbf{60.53} & Exos & 32 & 8 & 1 & 4.00 & High-frequency, external views \\
Dancing & 26.50 & Ego+Exos & 32 & 8 & 1 & 4.00 & High-frequency \\
Music & \textbf{74.14} & Ego & 32 & 12 & 3 & 0.89 &  Fine-grained, egocentric capture\\
Bouldering & \textbf{42.31} & Ego & 32 & 2 & 3 & 5.33 &  Rapid sampling, proprioceptive focus\\
Soccer & 66.67 & All & 24/32 & Various & 3 & Various & Consistent across configs \\
\bottomrule
\end{tabular}
\end{table*}

\begin{table*}[ht]
\centering
\caption{Accuracy comparison across different approaches and viewing configurations.}
\label{tab:performance_comparison}
\footnotesize
\begin{tabular}{lcccccccccc}
\toprule
\multirow{2}{*}{\textbf{Scenario}} & \multirow{2}{*}{\textbf{Maj.}} & \multicolumn{3}{c}{\textbf{Baseline}} & \multicolumn{3}{c}{\textbf{SkillFormer}} & \multicolumn{3}{c}{\textbf{SkillFormer+PATS}} \\
\cmidrule(lr){3-5} \cmidrule(lr){6-8} \cmidrule(lr){9-11}
& & \textbf{Ego} & \textbf{Exos} & \textbf{Ego+Exos} & \textbf{Ego} & \textbf{Exos} & \textbf{Ego+Exos} & \textbf{Ego} & \textbf{Exos} & \textbf{Ego+Exos} \\
\midrule
Basketball & 36.19 & 51.43 & 52.30 & 55.24 & 69.03 & 70.80 & \underline{77.88} & 64.60 & 72.57 & \textbf{78.76} \\
Cooking & 50.00 & 45.00 & 35.00 & 35.00 & 31.58 & 47.37 & \textbf{60.53} & 39.47 & \textbf{60.53} & \underline{50.11} \\
Dancing & \underline{51.61} & \textbf{55.65} & 42.74 & 42.74 & 20.51 & 15.38 & 13.68 & 22.22 & 20.51 & 26.50\\
Music & 58.97 & 46.15 & 69.23 & 56.41 & \underline{72.41} & 68.97 & 68.10 & \textbf{74.14} & 69.83 & 69.01 \\
Bouldering & 0.00 & 25.31 & 17.28 & 17.28 & 30.77 & 33.52 & 31.87 & \textbf{42.31} & \underline{36.81} & 36.26 \\
Soccer & 62.50 & 56.25 & \textbf{75.00} & \textbf{75.00} & \underline{70.83} & 66.67 & 66.67 & 66.67 & 66.67 & 66.67 \\
\bottomrule
\end{tabular}
\end{table*}

\section{Results}
\label{sec:results}
We evaluate PATS through three key dimensions: (1) overall accuracy against state-of-the-art baselines, (2) systematic analysis of PATS parameters, and (3) scenario-specific analysis. All experiments are conducted on the EgoExo4D benchmark under consistent evaluation protocols.

\subsection{State-of-the-Art Comparison}
\label{subsec:overall_performance}
Table~\ref{tab:baseline_comparison} demonstrates that SkillFormer+PATS surpasses the state-of-the-art accuracy across all viewing configurations. Our approach delivers consistent improvements over the original SkillFormer: 47.3\% accuracy for egocentric views (+3.05\%), 46.6\% for exocentric views (+0.65\%), and 48.0\% for combined views (+1.05\%). These gains are achieved while maintaining computational efficiency with 14-27M parameters and 4 training epochs.

\subsection{PATS Parameter Analysis}
\label{subsec:configuration_analysis}
Our systematic grid search across 12 major configurations (Table~\ref{tab:all_configurations}) reveals clear optimization patterns for proficiency estimation, synthesized in Table~\ref{tab:config_patterns}.

Three key principles emerge from the analysis. First, 32 frames proves universally optimal across all scenarios, while sampling rates diverge by activity type: dynamic activities require high rates (4.0-5.33 FPS) while structured sequential activities perform best at lower rates (0.89 FPS). Second, view selection aligns with skill characteristics: egocentric views excel for proprioceptive activities (bouldering, music), while fundamental-based sports benefit from fused ego-exocentric perspectives for comprehensive technique analysis. Third, temporal segmentation correlates inversely with action continuity—continuous actions require fewer segments (2) while fine-grained sequential skills benefit from finer segmentation (12).

These consistent patterns across diverse activities demonstrate PATS' adaptability to different skill assessment domains.

\subsection{Scenario-Specific Analysis}
\label{subsec:activity_performance}
Table~\ref{tab:best_configurations} shows optimal configurations for each scenario, revealing domain-specific preferences. Basketball achieves the highest accuracy (78.76\%) using rapid multi-view sampling (5.33 FPS, 2 segments), while music reaches 74.14\% through fine-grained egocentric capture (0.89 FPS, 12 segments). Cooking performs best with exocentric-only views (60.53\%) using high-frequency sampling, while bouldering achieves 42.31\% with rapid egocentric sampling.

Table~\ref{tab:performance_comparison} provides detailed per-scenario comparisons. PATS delivers substantial improvements in several domains: bouldering shows the largest gain (+26.22\% over SkillFormer), music improves by +2.39\%, and basketball by +1.13\%. These results demonstrate PATS' effectiveness for proprioceptive skills requiring precise temporal coordination.

However, PATS shows mixed results in some scenarios. For dancing, while PATS improves over SkillFormer (26.50\% vs 20.51\%), it remains below baseline methods (55.65\%), suggesting this domain may require alternative parameter combinations. For soccer, PATS shows a decline in egocentric view accuracy (66.67\% vs 70.83\%), indicating egocentric temporal sampling may be less suitable for soccer assessment.

Overall, these results demonstrate PATS' ability to adapt temporal sampling strategies to diverse skill domains while maintaining computational efficiency.

\section{Limitations and Future Work}
\label{sec:limitations}
Despite significant advances, PATS faces challenges in certain activity domains. In subjective domains like dancing, baseline methods outperform our approach, suggesting inadequate capture of important rhythmic and aesthetic components. Soccer presents another limitation, where PATS degrades egocentric performance while other viewing configurations remain stable, indicating that our sampling strategy may negatively impact specific activity-view combinations. The requirement for scenario-specific configuration also limits practical applicability across new domains.

Future research should focus on automated configuration selection mechanisms, enhanced temporal modeling for rhythmic activities, and multi-modal integration incorporating audio and haptic feedback. 

\section{Discussion and Conclusions}
\label{sec:discussion}
This work introduces PATS (Proficiency-Aware Temporal Sampling), achieving consistent state-of-the-art improvements across egocentric, exocentric, and combined viewing configurations (+0.65\% to +3.05\%) while maintaining computational efficiency. Our systematic analysis reveals actionable design principles: 32 frames prove universally optimal, sampling rates should match activity dynamics (high for dynamic, low for structured activities), and segmentation strategies must align with action continuity.

The substantial scenario-specific improvements, particularly in bouldering (+26.22\%), music (+2.39\%), and basketball (+1.13\%), validate that domain-aware temporal sampling meaningfully enhances proficiency estimation. The alignment between optimal configurations and activity semantics demonstrates PATS' semantic grounding and adaptability across diverse skill domains.

PATS represents a significant advancement in automated skill assessment, providing practitioners with principled temporal sampling guidelines while establishing a robust foundation for accurate and interpretable skill assessment systems in sports training and education.

\section*{Acknowledgements}
We acknowledge ISCRA for awarding this project access to the LEONARDO supercomputer, owned by the EuroHPC Joint Undertaking, hosted by CINECA (Italy).

\bibliographystyle{IEEEtran}
\bibliography{IEEEabrv,IEEEexample}

\begin{thebibliography}{10}
\providecommand{\url}[1]{#1}
\csname url@samestyle\endcsname
\providecommand{\newblock}{\relax}
\providecommand{\bibinfo}[2]{#2}
\providecommand{\BIBentrySTDinterwordspacing}{\spaceskip=0pt\relax}
\providecommand{\BIBentryALTinterwordstretchfactor}{4}
\providecommand{\BIBentryALTinterwordspacing}{\spaceskip=\fontdimen2\font plus
\BIBentryALTinterwordstretchfactor\fontdimen3\font minus \fontdimen4\font\relax}
\providecommand{\BIBforeignlanguage}[2]{{%
\expandafter\ifx\csname l@#1\endcsname\relax
\typeout{** WARNING: IEEEtran.bst: No hyphenation pattern has been}%
\typeout{** loaded for the language `#1'. Using the pattern for}%
\typeout{** the default language instead.}%
\else
\language=\csname l@#1\endcsname
\fi
#2}}
\providecommand{\BIBdecl}{\relax}
\BIBdecl

\bibitem{skillformer}
\BIBentryALTinterwordspacing
E.~Bianchi and A.~Liotta, ``Skillformer: Unified multi-view video understanding for proficiency estimation,'' 2025. [Online]. Available: \url{https://arxiv.org/abs/2505.08665}
\BIBentrySTDinterwordspacing

\bibitem{wang2019temporal}
L.~Wang, Y.~Xiong, Z.~Wang, Y.~Qiao, D.~Lin, X.~Tang, and L.~Van~Gool, ``Temporal segment networks for action recognition in videos,'' \emph{IEEE Transactions on Pattern Analysis and Machine Intelligence}, vol.~41, no.~11, pp. 2740--2755, 2019.

\bibitem{liu2025future}
Y.~Liu \emph{et~al.}, ``When the future becomes the past: Taming temporal correspondence for self-supervised video representation learning,'' \emph{arXiv preprint arXiv:2503.15096}, 2025.

\bibitem{kim2022capturing}
K.~Kim, S.~N. Gowda, O.~Mac~Aodha, and L.~Sevilla-Lara, ``Capturing temporal information in a single frame: Channel sampling strategies for action recognition,'' in \emph{33rd British Machine Vision Conference 2022, {BMVC} 2022}.\hskip 1em plus 0.5em minus 0.4em\relax BMVA Press, 2022.

\bibitem{kim2024leveraging}
M.~Kim, D.~Han, T.~Kim, and B.~Han, ``Leveraging temporal contextualization for video action recognition,'' in \emph{European Conference on Computer Vision}.\hskip 1em plus 0.5em minus 0.4em\relax Springer, 2024, pp. 74--91.

\bibitem{zhou2024comprehensive}
K.~Zhou, R.~Cai, L.~Wang, H.~P. Shum, and X.~Liang, ``A comprehensive survey of action quality assessment: Method and benchmark,'' \emph{arXiv preprint arXiv:2412.11149}, 2024.

\bibitem{carreira2017quo}
J.~Carreira and A.~Zisserman, ``Quo vadis, action recognition? a new model and the kinetics dataset,'' in \emph{Proceedings of the IEEE conference on computer vision and pattern recognition}, 2017, pp. 6299--6308.

\bibitem{liu2022video}
Z.~Liu, J.~Ning, Y.~Cao, Y.~Wei, Z.~Zhang, S.~Lin, and H.~Hu, ``Video swin transformer,'' in \emph{Proceedings of the IEEE/CVF conference on computer vision and pattern recognition}, 2022, pp. 3202--3211.

\bibitem{wang2021tsa}
S.~Wang, D.~Yang, P.~Zhai, C.~Chen, and L.~Zhang, ``Tsa-net: Tube self-attention network for action quality assessment,'' in \emph{Proceedings of the 29th ACM International Conference on Multimedia}, 2021, pp. 4902--4910.

\bibitem{xu2022finediving}
J.~Xu, Y.~Rao, X.~Yu, G.~Chen, J.~Zhou, and J.~Lu, ``Finediving: A fine-grained dataset for procedure-aware action quality assessment,'' in \emph{Proceedings of the IEEE/CVF Conference on Computer Vision and Pattern Recognition}, 2022, pp. 2949--2958.

\bibitem{tang2020uncertainty}
Y.~Tang, Z.~Ni, J.~Zhou, D.~Zhang, J.~Lu, Y.~Wu, and J.~Zhou, ``Uncertainty-aware score distribution learning for action quality assessment,'' in \emph{Proceedings of the IEEE/CVF Conference on Computer Vision and Pattern Recognition}, 2020, pp. 9839--9848.

\bibitem{yu2021group}
X.~Yu, Y.~Rao, W.~Zhao, J.~Lu, and J.~Zhou, ``Group-aware contrastive regression for action quality assessment,'' in \emph{Proceedings of the IEEE/CVF International Conference on Computer Vision}, 2021, pp. 7919--7928.

\bibitem{gsp}
E.~Bianchi and O.~Lanz, ``Gate-shift-pose: Enhancing action recognition in sports with skeleton information,'' in \emph{Proceedings of the Winter Conference on Applications of Computer Vision (WACV) Workshops}, February 2025, pp. 1257--1264.

\bibitem{skeletonAR}
H.~Duan, Y.~Zhao, K.~Chen, D.~Lin, and B.~Dai, ``Revisiting skeleton-based action recognition,'' in \emph{2022 IEEE/CVF Conference on Computer Vision and Pattern Recognition (CVPR)}, 2022, pp. 2959--2968.

\bibitem{skatinmixer}
\BIBentryALTinterwordspacing
J.~Xia, M.~Zhuge, T.~Geng, S.~Fan, Y.~Wei, Z.~He, and F.~Zheng, ``Skating-mixer: Long-term sport audio-visual modeling with mlps,'' 2022. [Online]. Available: \url{https://arxiv.org/abs/2203.03990}
\BIBentrySTDinterwordspacing

\bibitem{pamfn}
L.-A. Zeng and W.-S. Zheng, ``Multimodal action quality assessment,'' \emph{IEEE Transactions on Image Processing}, 2024.

\bibitem{timesformer}
\BIBentryALTinterwordspacing
G.~Bertasius, H.~Wang, and L.~Torresani, ``Is space-time attention all you need for video understanding?'' 2021. [Online]. Available: \url{https://arxiv.org/abs/2102.05095}
\BIBentrySTDinterwordspacing

\bibitem{lora}
\BIBentryALTinterwordspacing
E.~J. Hu, Y.~Shen, P.~Wallis, Z.~Allen-Zhu, Y.~Li, S.~Wang, L.~Wang, and W.~Chen, ``Lora: Low-rank adaptation of large language models,'' 2021. [Online]. Available: \url{https://arxiv.org/abs/2106.09685}
\BIBentrySTDinterwordspacing

\bibitem{egoppg}
\BIBentryALTinterwordspacing
B.~Braun, R.~Armani, M.~Meier, M.~Moebus, and C.~Holz, ``egoppg: Heart rate estimation from eye-tracking cameras in egocentric systems to benefit downstream vision tasks,'' 2025. [Online]. Available: \url{https://arxiv.org/abs/2502.20879}
\BIBentrySTDinterwordspacing

\bibitem{egoexolearn}
Y.~Huang, G.~Chen, J.~Xu, M.~Zhang, L.~Yang, B.~Pei, H.~Zhang, D.~Lu, Y.~Wang, L.~Wang, and Y.~Qiao, ``Egoexolearn: A dataset for bridging asynchronous ego- and exo-centric view of procedural activities in real world,'' in \emph{Proceedings of the IEEE/CVF Conference on Computer Vision and Pattern Recognition}, 2024.

\bibitem{egoexo4d}
K.~Grauman, A.~Westbury, L.~Torresani, K.~Kitani, J.~Malik, T.~Afouras, K.~Ashutosh, V.~Baiyya, S.~Bansal, B.~Boote, E.~Byrne, Z.~Chavis, J.~Chen, F.~Cheng, F.-J. Chu, S.~Crane, A.~Dasgupta, J.~Dong, M.~Escobar, C.~Forigua, A.~Gebreselasie, S.~Haresh, J.~Huang, M.~M. Islam, S.~Jain, R.~Khirodkar, D.~Kukreja, K.~J. Liang, J.-W. Liu, S.~Majumder, Y.~Mao, M.~Martin, E.~Mavroudi, T.~Nagarajan, F.~Ragusa, S.~K. Ramakrishnan, L.~Seminara, A.~Somayazulu, Y.~Song, S.~Su, Z.~Xue, E.~Zhang, J.~Zhang, A.~Castillo, C.~Chen, X.~Fu, R.~Furuta, C.~Gonzalez, P.~Gupta, J.~Hu, Y.~Huang, Y.~Huang, W.~Khoo, A.~Kumar, R.~Kuo, S.~Lakhavani, M.~Liu, M.~Luo, Z.~Luo, B.~Meredith, A.~Miller, O.~Oguntola, X.~Pan, P.~Peng, S.~Pramanick, M.~Ramazanova, F.~Ryan, W.~Shan, K.~Somasundaram, C.~Song, A.~Southerland, M.~Tateno, H.~Wang, Y.~Wang, T.~Yagi, M.~Yan, X.~Yang, Z.~Yu, S.~C. Zha, C.~Zhao, Z.~Zhao, Z.~Zhu, J.~Zhuo, P.~Arbelaez, G.~Bertasius, D.~Damen, J.~Engel, G.~M. Farinella, A.~Furnari, B.~Ghanem, J.~Hoffman, C.~Jawahar,
  R.~Newcombe, H.~S. Park, J.~M. Rehg, Y.~Sato, M.~Savva, J.~Shi, M.~Z. Shou, and M.~Wray, ``Ego-exo4d: Understanding skilled human activity from first- and third-person perspectives,'' in \emph{Proceedings of the IEEE/CVF Conference on Computer Vision and Pattern Recognition (CVPR)}, June 2024, pp. 19\,383--19\,400.

\bibitem{carreira17}
J.~Carreira and A.~Zisserman, ``Quo vadis, action recognition? a new model and the kinetics dataset,'' in \emph{2017 IEEE Conference on Computer Vision and Pattern Recognition (CVPR)}, 2017, pp. 4724--4733.

\end{thebibliography}
\end{document}